# Augmenting Text to Increase Translation Difficulty

**William Kalikman**[⋆] **Šimon Sukup**[⋆] **Michal Tešnar** **Vilém Zouhar**

ETH Zurich
{wkalikman,ssukup,mtesnar,vzouhar}@ethz.ch

## Abstract

As state-of-the-art machine translation models saturate standard benchmarks, the field needs more challenging evaluations to distinguish between models of varying quality. We propose augmenting existing benchmarks to increase translation difficulty by combining adversarial optimization with a differentiable translation difficulty estimator. Our **A**dversarial **T**ranslation **O**ptimization (ATO) uses gradients from a combined difficulty and fluency objective to iteratively replace tokens. Because each step branches over candidate substitutions at every position, optimization becomes a tree search problem, which we address with Beam Search. ATO offers a gradient-based alternative to LLM-based dataset creation without LLM prompting, expensive human curation, or task-specific model training. Our ATO-modified benchmark lowers average translation quality (xCOMET) from 0.93 to 0.82, compared to 0.88 for paraphrasing and 0.86 for a zero-shot baseline. Human evaluation shows the modified texts are somewhat less natural than the baselines but remain reasonably grammatical and plausible while being substantially harder to translate. We release two datasets of 350 English texts each[2], generated by our methods, as well as the code.[1]

## 1 Introduction

Recent advances in machine translation have led to performance saturation on standard benchmarks, with state-of-the-art models achieving near-identical scores (Kocmi et al., 2025; Akhtar et al., 2026). Distinguishing between models of varying quality requires more challenging evaluation data.

Existing efforts to build harder benchmarks follow three main strategies, each with distinct limitations. Expert-crafted challenge sets (Isabelle et al., 2017; Macketanz et al., 2018; Akhbardeh et al., 2021) are targeted but expensive and difficult to scale. Mining difficult texts from the web (Xu et al., 2025) is limited by the availability of naturally difficult content. LLM-based rewriting (Zouhar et al., 2025) delegates the task to a black-box model which exposes no per-position gradient indicating how each token contributes to difficulty or fluency. It therefore offers no mechanism to incorporate such constraints directly into the optimization.

We propose an alternative direction based on adversarial input optimization. Given a differentiable estimator of translation difficulty, we use its gradient signal to iteratively modify a source text so that it becomes harder to translate. We instantiate the difficulty estimator with Sentinel (Perrella et al., 2024), a regression model that predicts expected translation quality from source text alone. This reframes the generation of hard-to-translate text as finding adversarial examples that minimize the Sentinel score, connecting our work to existing methods for adversarial optimization in NLP. However, naively following the difficulty gradient degrades fluency: the optimizer finds texts that are hard to translate only because they are nonsensical. Generating text that is both difficult to translate and well-formed is therefore a constrained search problem. We address this by combining Greedy Coordinate Gradient-style (Zou et al., 2023) candidate selection with Beam Search and a differentiable grammaticality signal, jointly steering the optimization toward tokens that are difficult to translate and linguistically well-formed.

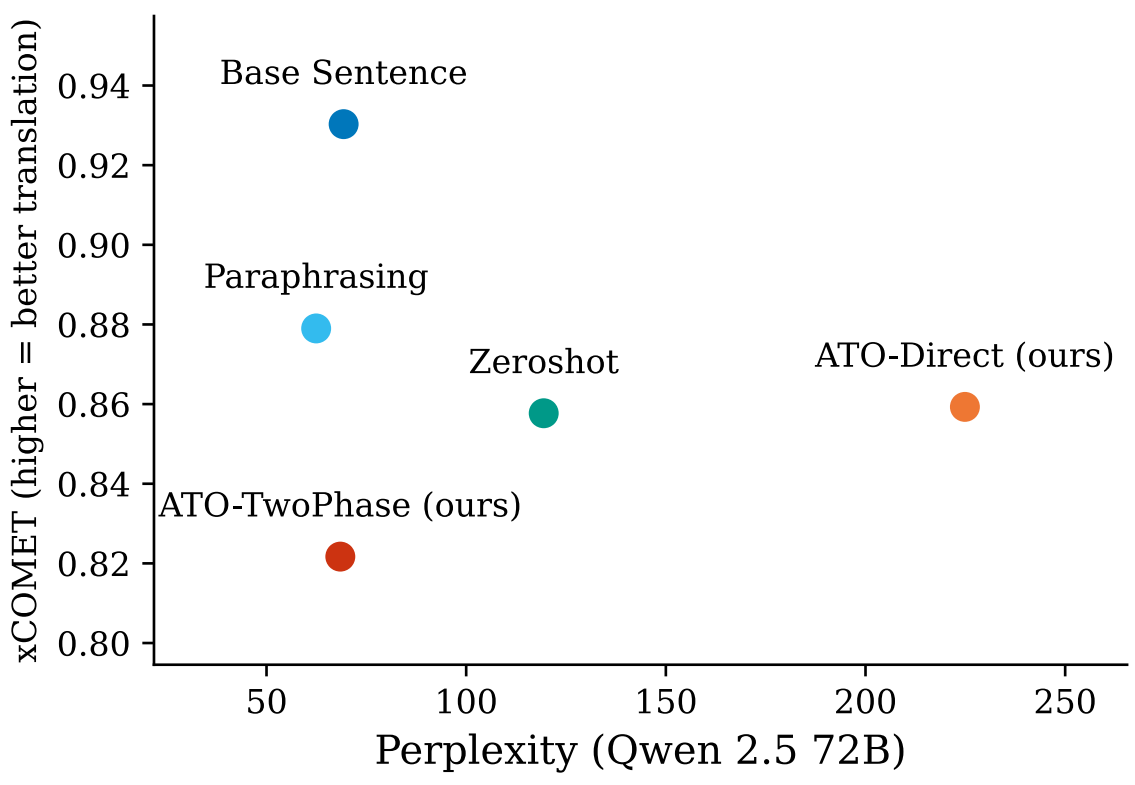

Figure 1: Text perplexity vs. translation quality (averaged across five languages and three models). Lower xCOMET indicates worse translation quality. ATO-TwoPhase reduces translation quality substantially while slightly decreasing perplexity.

| **Original:** *The Iraq Study Group presented its report at 12.00 GMT today.* | **Optimized:** *That odd old man insulted Marta at the market today.* |
|---|---|
| **Perplexity:** 19.77 | **Perplexity:** 70.11 |
| **MT Difficulty:** 53% | **MT Difficulty:** 71% |

Table 1: One example from our ATO-TwoPhase pipeline. The output is more difficult to translate while remaining fluent and grammatically correct.

[*]First authors with equal contribution
[2]huggingface.co/datasets/wskal/ATO-datasets
[1]github.com/BreakingMT/ATO

We call the resulting procedure Adversarial Translation Optimization (ATO) and present two variants. ATO-Direct restricts substitutions to whole-word tokens, enforcing fluency by construction and selecting the final output by perplexity, a standard proxy for fluency in text generation (Kann et al., 2018; Holtzman et al., 2020). ATO-TwoPhase first optimizes over the full subword vocabulary to generate hard-to-translate text in a first phase, then recovers fluency in a second phase by optimizing for perplexity under a different language model. Figure 1 and Table 1 illustrate this process: ATO-TwoPhase produces texts that are substantially harder to translate while maintaining fluency.

We evaluate our methods on 200 seed texts translated into five target languages by three translation models of varying capability. Both ATO variants produce harder-to-translate text than paraphrasing and zero-shot baselines: average xCOMET, a reference-free translation quality estimation metric, drops from 0.93 (base) to 0.82 (ATO-TwoPhase), compared to 0.88 for paraphrasing and 0.86 for the zero-shot baseline. Through a human annotation study involving 13 multilingual participants, we confirm that the resulting translations are of lower quality.

Our contributions are (1) Adversarial Translation Optimization (ATO), a gradient-based method for increasing translation benchmark difficulty without LLM prompting, human curation, or task-specific model training; and (2) two datasets of 350 augmented English texts each, produced by ATO-Direct and ATO-TwoPhase.

## 2 Related Work

**Adversarial optimization.** Adversarial text optimization has been approached from several angles. Reinforcement learning (Vijayaraghavan and Roy, 2020), generative adversarial networks (Ren et al., 2020), and continuous relaxation methods that operate in the embedding space (Ebrahimi et al., 2018; Sadrizadeh et al., 2023) have all been applied, but these approaches offer limited control over individual token-level substitutions. HotFlip (Ebrahimi et al., 2018) introduced the use of gradients with respect to one-hot token representations to identify single-token substitutions. Greedy Coordinate Gradient (Zou et al., 2023) computes top-k candidate replacements at every modifiable token position, then evaluates a random batch of single-token swaps drawn from across all positions, selecting the substitution that most reduces the adversarial loss. While effective at finding adversarial suffixes for LLM jailbreaking, it produces disfluent or nonsensical text, as the optimization has no incentive to preserve fluency or grammaticality.

**Fluency-constrained adversarial search.** Several methods address this fluency limitation. BeamAttack (Zhu et al., 2023) extends gradient-guided substitution with Beam Search, allowing exploration of locally suboptimal candidates that may lead to more fluent texts in later iterations. BESA (Yang et al., 2021) uses a masked language model to propose replacements that are simultaneously fluent and adversarially effective, combined with energy-based annealing to escape local minima. AutoDAN (Liu et al., 2024) departs from gradient-based methods, adopting a hierarchical genetic algorithm with LLM-based fitness metrics to maintain fluency. These methods demonstrate that balancing adversarial effectiveness with fluency is a recurring challenge, though all operate in the setting of adversarial suffix generation for LLM safety.

**Dataset curation and generation.** Our work also addresses limitations in how difficult translation benchmarks are sourced. Traditional challenge sets rely on expert curation (Kocmi et al., 2025), which is targeted but expensive and difficult to scale. Filtering approaches search existing web corpora for sentences that models naturally strug-gle with (Proietti et al., 2025; Xu et al., 2025), but are limited by the availability of naturally difficult text. More recently, LLM-based methods generate difficult texts through zero-shot or iterative prompting (Pombal et al., 2025; Zouhar et al., 2025), but as black-box generation, this approach yields no per-position gradient over the objective. This precludes gradient-based optimization, including approaches that combine multiple differentiable losses, such as our combination of difficulty and fluency.

## 3 Methods

Our work differs from previous adversarial optimization approaches in three ways. First, we optimize the full source text rather than appending adversarial suffixes, since our goal is to produce complete, fluent texts. Second, we target translation difficulty rather than LLM jailbreaking, using a learned translation difficulty estimator as part of the optimization objective. Third, rather than relying on post-hoc filtering for fluency, we incorporate a differentiable grammaticality signal directly into the gradient computation, jointly steering candidate selection toward tokens that are both difficult to translate and linguistically well-formed. Our search procedure combines Greedy Coordinate Gradient-style gradient-guided candidate selection with Beam Search to escape local minima and find fluent, difficult-to-translate sequences. We refer to the full procedure as Adversarial Translation Optimization (ATO).

### 3.1 Optimization Objective

We optimize over a token sequence $t = \langle t_1, ..., t_L \rangle$ composed of vocabulary tokens $t_i \in V$. Our goal is to maximize translation difficulty while maintaining fluency. We assume a differentiable scoring function $D(t)$ that estimates how difficult a source text $t$ is to translate, where lower values indicate greater difficulty. The unconstrained objective is:

$$t^* = \underset{t \in \mathcal{V}^L}{\arg\min}\ D(t) \tag{1}$$

We represent the text by a one-hot matrix $\boldsymbol{T} \in \{0,1\}^{L \times |\mathcal{V}|}$, where each row $\boldsymbol{T}_i$ encodes the token at position $i$. Because the one-hot representation enters the model through a differentiable embedding lookup, we can compute the gradient of the score with respect to the input:

$$\nabla_{\boldsymbol{T}} D = \left[ \frac{\partial D}{\partial \boldsymbol{T}_{i,v}} \right]_{i \in \{1,...,L\},\, v \in \mathcal{V}} \tag{2}$$

The entry $\nabla_{\boldsymbol{T}_{i,v}} D$ gives a first-order approximation of the change in $D$ if the token at position $i$ were replaced by vocabulary token $v$. These gradients guide the candidate selection procedure described next.

**Difficulty estimation.** We instantiate the function $D$ with Sentinel-src-25 (Proietti et al., 2025; Perrella et al., 2024), a source-only translation difficulty estimator. Sentinel-src is a regression model built on XLM-RoBERTa-large (Conneau et al., 2020), trained on human translation quality judgments to predict the expected quality of a text's translation from the source text alone. It assigns lower scores to texts whose translations tend to be worse, and is fully differentiable with respect to its input. See example Sentinel judgments:

$D$(“Good morning!”) = 0.46
$D$(“The eccentric aristocrat was not …”) = 0.34

Generally, $D$ could be instantiated by any differentiable difficulty estimator, for example as the expected quality score from a differentiable translation model composed with a differentiable quality estimation metric.

### 3.2 Beam Search with Greedy Coordinate Gradient

Rather than starting from random tokens, we seed the search with a real text $t_{\text{seed}}$. By initializing from well-formed text, each single-token substitution begins from a grammatical area in the space of all possible texts, so the resulting candidate is likely to remain close to fluent. This makes the fluency constraints described in the following sections effective: they need only prevent gradual drift from grammaticality, rather than recover it from scratch. The optimizer maintains a beam of $B$ candidate texts (we use $B = 50$). At each iteration, the following steps are applied identically across all candidates in the beam:

1. **Backward pass.** Token gradients $\nabla_{\boldsymbol{T}} \mathcal{L}$ are computed over the full $L \times |\mathcal{V}|$ one-hot input for every beam member, where $\mathcal{L}$ is the loss defined in Section 3.3 and Section 3.4.
2. **Top-$k$ shortlist.** The top $K = 200$ (position, token) pairs ranked by gradient magnitude are selected from the flattened $L \times |\mathcal{V}|$ gradient matrix, reducing the search space to the most promising single-token substitutions. Positions corresponding to punctuation tokens are excluded, and positions that were substituted in either of the two preceding iterations are frozen, preventing the optimizer from repeatedly cycling the same positions.

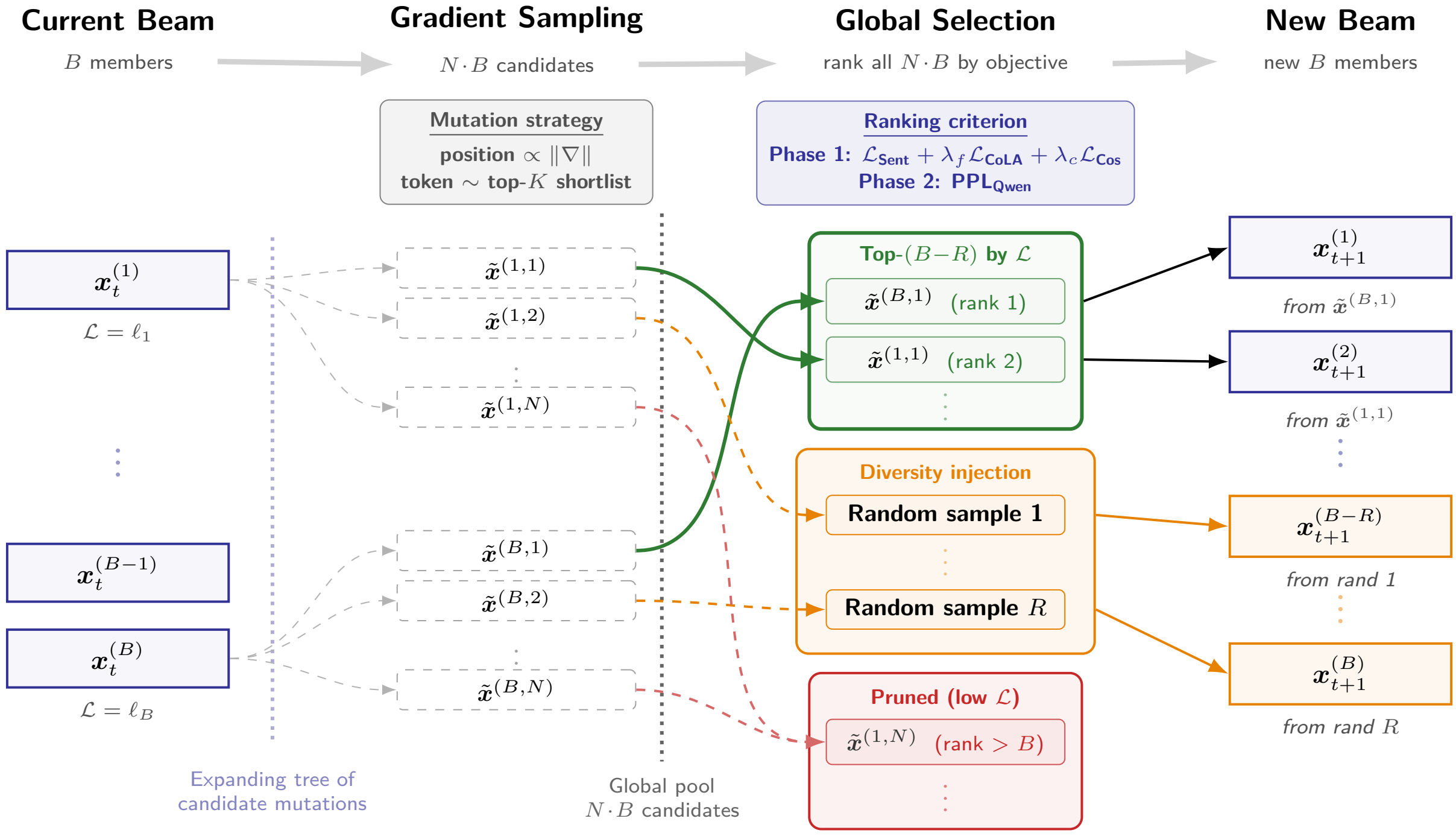

Figure 2: Beam search iteration visualization. In each iteration of ATO-Direct and ATO-TwoPhase, we first expand the current texts in the beam by testing candidate replacements under our constraints, prune based on the filtering metric, then choose the final set for the next beam. The random sampling of non-optimal beams ensures the beams do not collapse to a conforming set of texts due to local minima of the chosen objective.

3. **Pool sampling.** For each beam member, $N = 64$ candidate mutations are sampled: the position is drawn proportionally to its gradient norm, and the replacement token is drawn uniformly from that position's shortlist. This yields approximately $N \times B$ unique candidates per step.
4. **Objective scoring.** All candidates are evaluated under the current objective $\mathcal{L}$, defined in Section 3.3 and Section 3.4.
5. **Beam pruning.** The $B$ candidates with the lowest $\mathcal{L}$ are retained globally, with competitive selection across all beam parents.
6. **Diversity injection.** An additional 15 candidates are drawn randomly from the broader pool and added to the beam, preventing premature convergence.

The diagram of one iteration of Beam Search can be seen in Figure 2 and the corresponding algorithm in Algorithm 1. Left unconstrained, this optimization yields nonsensical text, which is difficult to translate only because it is ungrammatical, lacks semantic coherence, or contains fragments that do not form real words.

We address this with two schemes:

- ATO-Direct restricts the vocabulary to whole English words and selects the final output by choosing the lowest perplexity among candidates.
- ATO-TwoPhase allows the full subword vocabulary but follows the initial optimization with a second phase that explicitly optimizes for fluency.

See Table 2 for a comparative overview.

### 3.3 ATO-Direct

In ATO-Direct, one phase of Beam Search with Greedy Coordinate gradient is executed, optimizing only over whole-word tokens in the seed text with whole-word token candidates. For an overview, refer to Algorithm 2.

**Vocabulary.** We load a list of $\sim$0.5 million English words from the dwyl/english-words repository[2] and tokenize each with XLM-RoBERTa. We retain only words that the tokenizer encodes as a single token, yielding $\sim$10,000 to-

[2]github.com/dwyl/english-words

```
BeamSearchGCGStep(B_t, L(·), V, K, N, B, R):
1  for x ∈ B_t do G^(x) ← ∇_X L(x)          // Gradients measure token substitution impact
2  end for
3  S ← Top-K(G, B_t, V)                     // Shortlist S of most impactful tokens by gradient
4  P ← ∅
5  for x ∈ B_t do                           // initialize candidate pool P
6    for j = 1 to N do                      // N candidates in each beam member x
7      i* ~ p(i) ∝ ||G^(x)_{i,·}||          // Sample position i* by gradient norm
8      v* ~ Uniform(S_{i*})                 // Sample token v* from shortlist
9      P ← P ∪ {Replace(x, i*, v*)}         // Replace token and add candidate to P from shortlist
10   end for
11 end for
12 ∀ x̃ ∈ P : s(x̃) ← L(x̃)                   // Save ranking of candidates in mapping s
13 B_{t+1} ← Top_{B−R}(P, s)                // Top B-R candidates by objective score L
14 R ← SampleRandom(P \ B_{t+1}, R)         // R randomly sampled beam members for diversity injection
15 return B_{t+1} ∪ R, P                    // Returns B new members
```

Algorithm 1: One step of Beam Search. Parameters: Current Beam $\mathcal{B}_t$ of width $B$, objective $\mathcal{L}$, vocabulary $\mathcal{V}$, shortlist size $K$, pool samples $N$, random injections $R$.

kens. During optimization, positions in the seed text whose original word was split into multiple subword tokens are frozen; only positions that correspond to a single whole-word token are eligible for substitution. Replacements are likewise drawn exclusively from the ∼10,000 whole-word vocabulary, so every swap replaces one complete word with another.

**Objective.** To further encourage grammaticality beyond the vocabulary constraint, we introduce a differentiable fluency term by fine-tuning a grammatical acceptability classifier on the Corpus of Linguistic Acceptability (CoLA, Warstadt et al., 2019). We fine-tune XLM-RoBERTa-large (Conneau et al., 2020) for binary sequence classification (acceptable / unacceptable) using the CoLA training set from the GLUE benchmark (Wang et al., 2019). Sentinel-src also uses XLM-RoBERTa-large as its encoder, so both models share the same tokenizer and vocabulary. This allows us to compute gradients from both models with respect to a single one-hot representation $\boldsymbol{T}$.

The CoLA fluency loss is the cross-entropy between the classifier's output and the target label "acceptable":

$$\mathcal{L}_{\text{CoLA}}(t) = -\log P[\text{CoLA}(\text{acceptable} \mid t)] \quad (3)$$

To discourage beam candidates from fragmenting into dissimilar clusters, we introduce a cosine-similarity penalty over sentence-level representations. Following (Ebrahimi et al., 2018), we use the CLS-pooled representation, obtained from the XLM-RoBERTa backbone already employed in Sentinel. This term penalizes each candidate for drifting from the beam mean:

$$\mathcal{L}_{\cos}(t) = 1 - \cos\big(\boldsymbol{h}, \overline{\boldsymbol{h}}\big) \quad (4)$$

where $\boldsymbol{h}$ is the candidate's CLS representation and $\overline{\boldsymbol{h}}$ is the mean CLS representation across the current beam. The full objective combines all three terms:

$$\mathcal{L}_{\text{sent+CoLA}}(t) = \underbrace{D(t)}_{\text{difficulty}} + \lambda_f \cdot \underbrace{\mathcal{L}_{\text{CoLA}}(t)}_{\text{fluency}} + \lambda_c \cdot \underbrace{\mathcal{L}_{\cos}(t)}_{\text{cohesion}} \quad (5)$$

with $\lambda_f = 20$ and $\lambda_c = 50$. Because all three terms are differentiable with respect to $\boldsymbol{T}$, the gradient $\nabla_{\boldsymbol{T}} \mathcal{L}_{\text{sent+CoLA}}$ jointly steers candidate selection toward tokens that are difficult to translate, grammatically natural, and consistent across the beam. This is the objective $\mathcal{L}$ referenced in Section 3.2.

**Selection.** All candidates from all steps of the Beam Search are scored with Qwen 2.5-72B perplexity. The candidate with the lowest perplexity is selected as the final output.

### 3.4 ATO-TwoPhase

ATO-Direct's whole-word constraint enforces fluency but limits the search space. ATO-TwoPhase takes a two-phased approach to allow for a broader search: it allows the full subword vocabulary in a first phase of Beam Search to reach lower Sentinel scores, then cleans up the resulting text in a second Beam Search phase by optimizing for fluency.

**Phase 1.** Using the same ∼0.5 million word English list, we tokenize each word with XLM-RoBERTa and retain all tokens that appear in any tokenization. This yields ∼25,000 tokens: the vocabulary now includes subword fragments but remains limited to English.

ATO-DIRECT$(\boldsymbol{x}_{1:L}, \mathcal{V}_{\text{word}}, T, K, N, B, R)$:

1 $\mathcal{B}_0 \leftarrow \{\boldsymbol{x}_{1:L}\}$ // Seed beam with input sentence
2 $|(\boldsymbol{h}) \leftarrow \mathrm{CLS}(\boldsymbol{x}_{1:L})$ // Initial beam-mean embedding
3 $\mathcal{C} \leftarrow \varnothing$ // Global candidate archive
4 **for** $t = 1$ **to** $T$ **do**
5 $\quad \mathcal{L}_1(\cdot) \leftarrow \mathcal{L}_{\text{Sentinel}} + \lambda_f \mathcal{L}_{\text{CoLA}} + \lambda_c(1 - \cos(\mathrm{CLS}(\cdot), |(\boldsymbol{h})))$
6 $\quad \mathcal{B}_t, \mathcal{P}_t \leftarrow$ BEAMSEARCHSTEP $(\mathcal{B}_{t-1}, \mathcal{L}_1, \mathcal{V}_{\text{word}}, K, N, B, R)$
7 $\quad |(\boldsymbol{h}) \leftarrow \frac{1}{|}\mathcal{B}_t| \sum_{\boldsymbol{x}\in\mathcal{B}_t} \mathrm{CLS}(\boldsymbol{x})$ // Update beam-mean embedding
8 $\quad \mathcal{C} \leftarrow \mathcal{C} \cup \mathcal{P}_t$ // Archive all candidates
9 **end for**
10 $\boldsymbol{x}^\star \leftarrow \arg\min_{\boldsymbol{x}\in\mathcal{C}} \mathrm{PPL}_{\mathrm{Qwen}(\boldsymbol{x})}$ // Save lowest-PPL text
11 **return** $\boldsymbol{x}^\star$

Algorithm 2: In ATO-Direct, the single Beam Search Greedy Coordinate Gradient phase uses the same composite objective as ATO-TwoPhase but restricts substitutions to complete English words ($|\mathcal{V}_{\text{word}}| \approx$10k), enforcing fluency by construction. Selection is done by Qwen 2.5-72B perplexity ranking over all candidates across all steps.

ATO-TWOPHASE$(\boldsymbol{x}_{1:L}, \mathcal{V}_{\text{sub}}, \mathcal{V}_{\text{Qwen}}, T_1, T_2, K, N, B, R)$:

1 $\mathcal{B}_0 \leftarrow \{\boldsymbol{x}_{1:L}\}$ // Seed beam with input text
2 $|(\boldsymbol{h}) \leftarrow \mathrm{CLS}(\boldsymbol{x}_{1:L})$ // Initial beam-mean embedding
3 **for** $t = 1$ **to** $T_1$ **do**
4 $\quad \mathcal{L}_1(\cdot) \leftarrow \mathcal{L}_{\text{Sentinel}} + \lambda_f \mathcal{L}_{\text{CoLA}} + \lambda_c(1 - \cos(\mathrm{CLS}(\cdot), |(\boldsymbol{h})))$
5 $\quad \mathcal{B}_t, \mathcal{P}_t \leftarrow$ BEAMSEARCHSTEP $(\mathcal{B}_{t-1}, \mathcal{L}_1, \mathcal{V}_{\text{sub}}, K, N, B, R)$
6 $\quad |(\boldsymbol{h}) \leftarrow \frac{1}{|}\mathcal{B}_t| \sum_{\boldsymbol{x}\in\mathcal{B}_t} \mathrm{CLS}(\boldsymbol{x})$ // Update beam-mean embedding
7 **end for**
8 $\mathcal{B}_{T_1} \leftarrow \text{Re-tokenise}\big(\mathcal{B}_{T_1}, \mathcal{V}_{\text{Qwen}}\big)$ // Switch to Qwen token space
9 $\mathcal{C} \leftarrow \varnothing$ // Global candidate archive
10 **for** $t = T_1 + 1$ **to** $T_1 + T_2$ **do**
11 $\quad \mathcal{B}_t, \mathcal{P}_t \leftarrow$ BEAMSEARCHSTEP $(\mathcal{B}_{t-1}, \mathrm{PPL}_{\text{Qwen}}, \mathcal{V}_{\text{Qwen}}, K, N, B, R)$
12 $\quad \mathcal{C} \leftarrow \mathcal{C} \cup \mathcal{P}_t$ // Archive all candidates
13 **end for**
14 $\boldsymbol{x}^\star \leftarrow \arg\min_{\boldsymbol{x}\in\mathcal{C}} \mathrm{PPL}_{\mathrm{Qwen}(\boldsymbol{x})}$ // Lowest PPL across all Phase-2 steps
15 **return** $\boldsymbol{x}^\star$

Algorithm 3: ATO-TwoPhase. Phase 1 optimizes Sentinel difficulty, CoLA fluency, and cosine similarity over the full XLM-R subword vocabulary ($|\mathcal{V}_{\text{sub}}| \approx 25,000$). Phase 2 switches to Qwen's vocabulary ($|\mathcal{V}_{\text{Qwen}}| \approx 87,000$) and optimizes for perplexity. The output is the text with the lowest Qwen perplexity encountered across all Phase 2 candidates.

The objective is identical to ATO-Direct (Equation 5): $\mathcal{L}_{\text{sent+CoLA}}$, combining Sentinel difficulty, CoLA fluency, and cosine cohesion with the same hyperparameters. The only difference is that every token in the original text can be changed, unlike in Section 3.3 where only whole-word tokens were candidates for optimization. Because the vocabulary is more permissive, the optimizer can reach lower Sentinel scores, but may produce texts containing combinations of subword fragments that do not form coherent words.

After the Beam Search completes, the single candidate with the lowest Sentinel score (highest estimated translation difficulty) is selected to seed Phase 2.

**Phase 2** operates in Qwen 2.5′s token space. We apply an ASCII filter over Qwen's full vocabulary (∼150,000 tokens), retaining any token whose decoded string contains only ASCII letters, digits, spaces, and common punctuation. This yields ∼87,000 tokens: English-compatible subword fragments with non-Latin scripts removed.

The Phase 1 output is replicated into a fresh beam of width $B = 50$. The Beam Search then runs for 40 steps using Qwen 2.5-72B perplexity as the sole objective:

$$\mathcal{L}_{\text{PPL}}(t) = \mathrm{PPL}_{\mathrm{Qwen}(t)} \tag{6}$$

Gradients now flow through Qwen's embedding matrix rather than XLM-RoBERTa's. The optimizer structure is otherwise identical to Phase 1.

The candidate with the lowest perplexity across all Phase 2 steps is selected as the final output. See Algorithm 3 for the method overview.

### 3.5 Baselines

We compare our methods against three baselines that a practitioner would intuitively use to generate harder-to-translate text.

**Zeroshot.** Our ATO-Direct approach replaces one to three words of the original text. As a comparable baseline, we prompt two LLMs to replace exactly two words in a text to make it harder to translate (see prompt in Figure 4): Qwen2.5-72B-Instruct, which doubles as the perplexity scorer in ATO, and the more recent frontier model DeepSeek-V4-Flash (DeepSeek-AI, 2026). See Appendix A for per-language results.

| | ATO-Direct | ATO-TwoPhase — Phase 1 | ATO-TwoPhase — Phase 2 |
|---|---|---|---|
| Vocabulary | Whole-word (9,865) | English subwords (25,329) | Qwen English (87,221) |
| Eligible positions | Single-token words only | All positions | All positions |
| Objective | $\mathcal{L}_{\text{sent+CoLA}}$ | $\mathcal{L}_{\text{sent+CoLA}}$ | $\mathcal{L}_{\text{PPL}}$ |
| Selection | Lowest Qwen PPL | Lowest Sentinel score | Lowest Qwen PPL |
| Fluency | Hard (vocabulary) + soft (CoLA) | Soft (CoLA) | Soft (PPL optimization) |

Table 2: Comparison of Greedy Coordinate Gradient Beam Search variants across our two ATO methods.

**Paraphrasing.** Our ATO-TwoPhase approach has a broader and less predictable effect, changing a variable number of words while preserving some semantic content. The paraphrasing model DIPPER (Krishna et al., 2023) is a comparable baseline, taking a text as input and outputting a rephrased version. We discuss the details of our DIPPER implementation in Appendix B.3.

**Random replacement.** To check our methods against random perturbations, we replace random tokens in each seed text; see Appendix A for implementation details and results.

## 4 Experiments

We evaluate ATO's ability to generate source texts that are harder to translate while preserving linguistic quality. Since our generated texts have no reference translations, we cannot use standard reference-based metrics such as BLEU (Papineni et al., 2002; Post, 2018). And since Sentinel-src is the optimization objective itself, using it to evaluate difficulty would be circular. We therefore assess difficulty by translating the generated texts with multiple models and measuring translation quality using reference-free automatic metrics and human judgments.

**Setup.** We evaluate on seed texts sampled from FLORES200 (NLLB Team, 2024), WMT22, WMT23, and WMT24 (Kocmi et al., 2022; Kocmi et al., 2023; Kocmi et al., 2024): 200 base texts for our automatic evaluation and 25 for our human evaluation. Each text is obtained by taking the first 50 tokens of a sample and truncating at the first sentence-ending punctuation; samples with no such punctuation are discarded. The resulting segments are typically 10–20 words long, and may be sentence fragments, titles, or full sentences depending on the source data. For each seed, we compare five variants in the main analysis: the original Base text, two baselines (DIPPER Paraphrasing and Qwen Zeroshot), and the two versions of our method (ATO-Direct and ATO-TwoPhase). The Random and DeepSeek-Zeroshot baselines are reported alongside in the appendix.

Each variant is translated from English to five target languages spanning several language families and resource levels: German, Spanish, Russian, Czech, and Icelandic. We use three translation models of varying capability: the encoder-decoder model NLLB-200-3.3B (NLLB Team, 2024), the open-source TranslateGemma-27b (Finkelstein et al., 2026), and the closed-source frontier LLM Gemini-3-Flash (Google DeepMind, 2025) see Figure 3 for the LLM translation prompt. We additionally compare against Tower-Plus-72B (Rei et al., 2025) and DeepSeek-V4-Flash (DeepSeek-AI, 2026) in Appendix A.

### 4.1 Automatic Evaluation

We use two standard reference-free metrics:

- **MetricX** (Juraska et al., 2024) estimates translation error on a scale from 0 (perfect) to 25 (worst). Higher scores indicate lower quality translations. For comparability, we report MetricX in Table 3 as $1 - \frac{\text{MetricX}}{25}$.
- **xCOMET** (Guerreiro et al., 2024) scores translations from 0 to 1, where 1 indicates a perfect translation. Lower scores indicate lower quality translations.

**Sentinel and perplexity.** Table 3 shows that both ATO variants substantially reduce Sentinel-src scores relative to the base texts and both baselines, confirming that the optimization successfully finds texts the difficulty estimator considers harder to translate. ATO-TwoPhase achieves the lowest Sen-tinel scores, as expected from its larger search space. Perplexity increases moderately for ATO-Direct but *decreases* for ATO-TwoPhase, reflecting Phase 2′s explicit perplexity optimization; the resulting texts are, by this measure, no less fluent than the originals. The Paraphrasing and Zeroshot baselines leave Sentinel scores approximately unchanged.

| Method | Perplexity | Sentinel | xCOMET | MetricX | (human) Grammatic. | (human) Plausible | (human) Transl. Qual. |
|---|---|---|---|---|---|---|---|
| Base text | 69.3 $\pm$42.62 | 0.29 $\pm$0.03 | 0.93 $\pm$0.02 | 0.91 $\pm$0.03 | 4.0 $\pm$0.28 | 4.4 $\pm$0.20 | 3.8 $\pm$0.31 |
| Paraphrasing | 62.4 $\pm$42.42 | 0.25 $\pm$0.03 | 0.88 $\pm$0.04 | 0.88 $\pm$0.04 | 4.0 $\pm$0.28 | 4.4 $\pm$0.19 | 3.9 $\pm$0.29 |
| Zeroshot (Qwen) | 119.4 $\pm$65.89 | 0.18 $\pm$0.03 | 0.86 $\pm$0.04 | 0.87 $\pm$0.05 | 3.8 $\pm$0.30 | 4.0 $\pm$0.23 | 3.9 $\pm$0.29 |
| ATO-Direct | 224.9 $\pm$95.25 | 0.02 $\pm$0.04 | 0.86 $\pm$0.04 | 0.85 $\pm$0.04 | 2.9 $\pm$0.30 | 3.0 $\pm$0.26 | 3.5 $\pm$0.32 |
| ATO-TwoPhase | 68.5 $\pm$11.95 | 0.00 $\pm$0.04 | 0.82 $\pm$0.03 | 0.82 $\pm$0.04 | 2.4 $\pm$0.27 | 2.8 $\pm$0.28 | 3.4 $\pm$0.28 |

Table 3: Aggregate evaluation scores comparing proposed ATO methods against baselines. Greener means better under the metrics we evaluated: higher fluency text, worse translation quality. MetricX scores are presented as $1 - \frac{\text{MetricX}}{25}$ for comparability. All values are presented as mean with 95% confidence interval.

**Translation quality.** Since Sentinel-src is the optimization objective itself, using it to evaluate difficulty would be circular. The key question is whether lower Sentinel scores correspond to genuinely worse translations under independent metrics. Table 3 confirms this across both metrics, with scores averaged across all five target languages and three translation models. On xCOMET, ATO-TwoPhase lowers scores from 0.93 to 0.82, outperforming both the Paraphrasing baseline (0.88) and Zeroshot (0.86). ATO-Direct performs comparably to Zeroshot on xCOMET (both 0.86) and MetricX (0.85 vs. 0.87 with overlapping CIs). On both metrics, ATO-TwoPhase achieves the largest quality decrease, and its differences from both baselines exceed the 95% confidence intervals.

**Variation across models.** The difficulty increase is consistent across all three translation models. NLLB-200-3.3B, the weakest model, shows the largest absolute xCOMET drop under ATO-TwoPhase (0.13 points averaged across languages), but the effect is not limited to weaker models: Gemini-3-Flash drops by 0.11 points and TranslateGemma by 0.09. This confirms that ATO-generated texts challenge models across capability levels, not only those that are already fragile. Full per-language, per-model breakdowns appear in Appendix A.

**Variation across languages.** The effect holds for all five target languages but varies in magnitude. Czech, Russian, and Icelandic show the largest xCOMET drops (0.12–0.14 points), while Spanish (0.10) and German (0.07) prove more resilient. See Appendix A for more details.

**ATO-Direct vs. -TwoPhase.** ATO-TwoPhase consistently achieves lower translation quality scores than ATO-Direct across all model–language pairs. As Figure 1 illustrates, ATO-TwoPhase also achieves lower perplexity thanks to its explicit fluency optimization phase. However, as we show in the human evaluation below, ATO-Direct's whole-word constraint produces texts that human raters judge as more grammatical and plausible than ATO-TwoPhase, suggesting that perplexity alone does not capture all aspects of naturalness.

## 4.2 Human Evaluation

We implemented two human evaluation schemes: one evaluating the quality of the generated English texts and another evaluating the quality of their translations. We recruited 13 annotators from the authors' academic network.

For English text quality, the respondents evaluated 125 total texts: five variants (the same as described in Section 4) of 25 base texts. Evaluators rated each text on a scale of 1–5 along two dimensions: *grammaticality*, how grammatically well-formed the text is, and *plausibility*, how likely it is that the text would appear in real online content. See the full evaluation instructions in Figure 5.

For translation quality, we translated all five variants of the 25 source texts into five target languages: German, Spanish, Russian, Czech, and Icelandic. We translated the texts with NLLB-200-3.3B. Annotators then rated each translation on a scale of 1–5. See the full transla-

**Phase 1**
*The Iraq Study Group presented its report at 12.00 GMT today.*
*The Iraq Study Group* ***banad*** *its report at 12.00 GMT today.*
*The Iraq Study Group* ***banad*** *its* ***duo*** *at 12.00 GMT today.*
*The Iraq Study Group* ***magasind*** *its* ***mister*** *at* ***off*** *GMT today.*
*The* ***flavor*** *Study Group* ***banad*** *its* ***duo*** *at* ***kul*** *GMT today.*
*The* ***sensitive shopping duo frontd*** *its* ***plate*** *at* ***leaving*** *GMT today.*
***bone*** *sensitive shopping duo frontd its* ***maya*** *at* ***keeping*** *GMT today.*

**Phase 2**
*bone sensitive shopping duo frontd its maya at keeping GMT today.*
*bone* ***bones*** *shopping duo frontd its maya at keeping GMT today.*
*bone* ***sensitive*** *shopping duo frontd its* ***npa*** *at* ***around*** *GMT today.*
*...*
***Some other*** *shopping clerk insulted* ***anna*** *at the supermarket* ***today.***
***That old shop*** *clerk insulted* ***my nona*** *at the supermarket* ***recently.***
***some other food*** *clerk insulted* ***anna*** *at the supermarket* ***today.***
*...*
*That old store clerk insulted my* ***grandpa*** *at the* ***river*** *market.*
***“s*** *old* ***shop*** *clerk insulted* ***his*** *grandpa at the* ***fish*** *market.*
***Quite*** *old shop* ***worker*** *insulted* ***my Grandpa*** *at the* ***local*** *market.*
*Quite old shop worker insulted my Grandpa at the local market.*
***Some other*** *old* ***woman*** *insulted* ***Marta*** *at the* ***market today.***
*Some other old woman insulted Marta at the market today.*
*...*
***“Oh that*** *old woman insulted* ***our*** *Marta at the market today.*
*“Oh that old woman insulted our Marta at the market today.*
***That odd*** *old* ***man*** *insulted Marta at the market today.*

Table 4: ATO-TwoPhase trace illustrated for one seed text. The fluent seed text is iteratively degraded by Phase 1, yielding a nonsensical but hard-to-translate segment. Phase 2 iteratively makes the Phase 1 output more fluent, resulting in a coherent, still hard-to-translate output.

tion evaluation instructions in Figure 6 and inter-annotator agreement in Appendix C.3. Screenshots of the evaluation interface can be seen in Appendix C.4.

Table 3 presents human ratings across the five methods with 95% confidence intervals. Both ATO-Direct and ATO-TwoPhase achieve lower translation quality scores compared to the baselines, confirming that ATO-generated texts are harder to translate. This comes at the cost of lower grammaticality and plausibility ratings compared to the baselines and seed texts. Between the two ATO variants, ATO-Direct achieves a comparable reduction in translation quality while retaining higher human-rated fluency, making it a more suitable choice for generating natural-sounding texts under the human evaluation metrics.

## 5 Discussion

The results confirm that ATO successfully lowers the translation quality compared to zero-shot and paraphrasing baselines. Across all automatic metrics and human judgments of translation quality, both ATO variants consistently produce harder-to-translate texts than either baseline, with ATO-TwoPhase achieving the largest difficulty increase. Nonetheless, the human evaluation Table 3 has shown a tradeoff between our naturalness proxies, grammaticality and plausibility, and the translation quality.

To illustrate how the two ATO-TwoPhase phases interact, we trace a single ATO-TwoPhase example from seed text to final output. Table 1 shows the input/output pair with its perplexity and MT difficulty scores, and Table 4 shows the full optimization trace, with bolded words marking changes at each step.

This example illustrates the complementary roles of the two phases. In Phase 1, individual tokens are replaced to maximize translation difficulty, often introducing ungrammatical or nonsensical words. The output of Phase 1 scores very low on the sentinel difficulty metric but is not human-readable. In Phase 2, the optimization objective shifts to perplexity, iteratively replacing tokens to recover fluency while implicitly preserving difficulty. The resulting text is fluent and grammatical, and as shown by the metrics in Table 1, remains more difficult to translate than the base text.

**Qualitative evaluation.** To better understand how ATO breaks machine translation, we examine several examples translated by NLLB-200-3.3B. A typical failure mode for ATO-Direct translations is the omission by the translation model of a contextually unlikely word that could nonetheless be translated directly. As shown in Figure 1, ATO-Direct produces texts with high perplexity, meaning they contain unexpected word combinations. Crucially, this does not mean the texts are untranslatable. In the first example in Table 5, the newly introduced word *run* could be translated as an adjective and included in the compound noun *die Laufinteressen*, yet NLLB omits it entirely.

Another failure mode of NLLB consists of failing to translate uncommon phrases. In the second example in Table 5, ATO-Direct introduces the awkward phrase *it was my cable to assist*, where NLLB centers its translation on the word *cable*, rendering it as *telegram* (*byl to můj telegram, abych vám pomohl*) rather than preserving the syntactic structure of the original text. In the third example, ATO-Direct introduces the phrase *sorry investment*, which could reasonably be interpreted literally as a *bad investment*. Instead, NLLB extrapolates beyond the source text and produces *inversión que inicialmente parecía un fracaso* (*investment that initially seemed like a failure*), adding meaning not present in the original text.

ATO-TwoPhase challenges translation in a different way. In the example in Table 6, the method produces the phrase *methods mince to people*. NLLB generates an incorrect translation, rendering the phrase as *nicht hilfreich sind*, *not being helpful* in German. The model appears to fill in a meaning that fits the surrounding context rather than faithfully translating the source, producing a fluent but incorrect output. Similarly to the first example in Table 5, we observe that a word out of context can worsen the model's performance. In the second example in Table 6, we see that the word *joist* is completely ignored. Finally, ATO can cause the translation model to drop content entirely: in the third example of Table 6, NLLB omits a full sentence that was translated correctly from the unmodified source. We hypothesize that the modified tokens push the encoder representations sufficiently out of distribution that the decoder's attention skips over the affected segment altogether.

| Augmented Text | Translation by NLLB |
|---|---|
| *We agree with the USOC's statement that **run** interests of our athletes and clubs, and their sport, may be better served by moving forward with meaningful change within our organization, rather than decertification.* | *Wir stimmen mit der Aussage des USOC überein, dass die Interessen unserer Athleten und Vereine und ihres Sports besser bedient werden können, wenn wir mit sinnvollen Veränderungen innerhalb unserer Organisation voranschreiten, anstatt die Zertifizierung zu entziehen.* |
| *Thank you for contacting #PRS_#ORG, it was my **cable** to assist you today.* | *Děkuji, že jste kontaktovali #PRS_ORG#, byl to můj telegram, abych vám dnes pomohl.* |
| *A **sorry** investment at launch would be worth over $2 million today!* | *¡Una inversión que inicialmente parecía un fracaso hoy valdría más de 2 millones de dólares!* |

Table 5: Low-quality translations by NLLB of texts produced by ATO-Direct, showcasing failure modes such as omitting contextually unexpected words and hallucinating explanatory phrases not present in the source.

| Augmented Text | Translation by NLLB |
|---|---|
| *With some other regularity, he is skeptical about how diabetes can be cured, noting that these methods will not **mince** to people who already have Type 1 diabetes.* | *Mit einer anderen Regelmäßigkeit ist er skeptisch, wie Diabetes geheilt werden kann, und stellt fest, dass diese Methoden für Menschen, die bereits Typ-1-Diabetes haben, nicht hilfreich sind.* |
| *The pump head and clamps are both covered by a lifetime manufacturer **joist**, so you can immediately replace the product in the event of shrinkage or spooling.* | *Hlavice čerpadla a svorky jsou oba kryty životním výrobcem, takže můžete okamžitě vyměnit výrobek v případě zmenšení nebo spoolingu.* |
| *His girlfriend outside of the video footage laughed as she took the video. Two Dollywood employees tried to break up the fight, including a manager pulling the two apart.* | *Dos empleados de Dollywood intentaron romper la pelea, incluido un gerente que los separó.* |

Table 6: Low-quality translations by NLLB of texts produced by ATO-TwoPhase, showcasing failure modes such as substituting contextually plausible but incorrect meanings, ignoring out-of-context words entirely, and dropping full sentences.

## 6 Conclusion

We introduced Adversarial Translation Optimization (ATO), a gradient-based method for augmenting text to increase translation difficulty. By combining Greedy Coordinate Gradient with Beam Search and a differentiable fluency signal, ATO iteratively modifies source texts to minimize Sentinel-src scores while preserving grammaticality. We presented two variants: ATO-Direct, which restricts substitutions to whole-word tokens, and ATO-TwoPhase, which operates over subword vocabularies to reach lower difficulty scores before recovering fluency through perplexity-guided optimization.

Experiments on 200 seed texts translated into five languages by three models of varying capability show that both ATO variants produce substantially harder-to-translate text than paraphrasing and zero-shot baselines, as measured by MetricX and xCOMET. Human evaluators confirm that the resulting translations are of lower quality, though ATO-generated texts were rated less grammatical and plausible than baselines. Future work should explore stronger fluency constraints or alternative optimization objectives to better reconcile translation difficulty with naturalness.

Our approach requires no LLM prompting, no human intervention or hand-crafted datasets, and provides per-token gradient signal that makes the optimization interpretable and enables direct enforcement of fluency constraints. ATO can be applied to any source text given a differentiable difficulty estimator, enabling scalable construction of challenging translation benchmarks. We release two datasets of 350 augmented texts, one produced by ATO-Direct and one produced by ATO-TwoPhase.

## Sustainability Statement

**Dataset creation.** Across both datasets, the total Phase 1 runtime was ∼10 hours on an RTX Pro 6000 GPU. Phase 2 runtime was ∼95.5 wall-clock hours across 2x RTX Pro 6000 GPUs. Qwen 72B perplexity scoring took ∼3 wall-clock hours on 2x RTX Pro 6000 GPUs, bringing the total to ∼204 RTX Pro 6000 GPU hours.

**Evaluation.** The translations we ran locally took ∼2 hours on a GeForce RTX 3090. We cannot provide a reliable estimate of the impact of our API usage for models accessed via API; however, we expect it to account for a very small share of total impact relative to dataset creation.

All experiments were run on private infrastructure in Switzerland. Accordingly, we use an electricity emissions factor of 0.09 kgCO2e/kWh per the latest reports on Swiss electricity consumption emissions intensity. We use the Machine Learning CO2 Impact Calculator, approximating our RTX Pro 6000 usage with an RTX A6000, as it is the most similar GPU available on the site. We calculate a total of 5.51 kg (dataset creation) + 0.06 kg (evaluation) ≈ 5.57 kg CO2.

## References

Farhad Akhbardeh, Arkady Arkhangorodsky, Magdalena Biesialska, Ondřej Bojar, Rajen Chatterjee, Vishrav Chaudhary, Marta R. Costa-jussa, Cristina España-Bonet, Angela Fan, Christian Federmann, Markus Freitag, Yvette Graham, Roman Grundkiewicz, Barry Haddow, Leonie Harter, Kenneth Heafield, Christopher Homan, Matthias Huck, Kwabena Amponsah-Kaakyire, Jungo Kasai, Daniel Khashabi, Kevin Knight, Tom Kocmi, Philipp Koehn, Nicholas Lourie, Christof Monz, Makoto Morishita, Masaaki Nagata, Ajay Nagesh, Toshiaki Nakazawa, Matteo Negri, Santanu Pal, Allahsera Auguste Tapo, Marco Turchi, Valentin Vydrin, and Marcos Zampieri. 2021. Findings of the 2021 Conference on Machine Translation (WMT21). In *Proceedings of the Sixth Conference on Machine Translation*, pages 1–88, Online.

Mubashara Akhtar, Anka Reuel, Prajna Soni, Sanchit Ahuja, Pawan Sasanka Ammanamanchi, Ruchit Rawal, Vilém Zouhar, Srishti Yadav, Chenxi Whitehouse, Dayeon Ki, Jennifer Mickel, Leshem Choshen, Marek Šuppa, Jan Batzner, Jenny Chim, Jeba Sania, Yanan Long, Hossein A. Rahmani, Christina Knight, Yiyang Nan, Jyoutir Raj, Yu Fan, Shubham Singh, Subramanyam Sahoo, Eliya Habba, Usman Gohar, Siddhesh Pawar, Robert Scholz, Arjun Subramonian, Jingwei Ni, Mykel Kochenderfer, Sanmi Koyejo, Mrinmaya Sachan, Stella Biderman, Zeerak Talat, Avijit Ghosh, and Irene Solaiman. 2026. When AI Benchmarks Plateau: A Systematic Study of Benchmark Saturation. arXiv: 2602.16763 [cs.AI].

Alexis Conneau, Kartikay Khandelwal, Naman Goyal, Vishrav Chaudhary, Guillaume Wenzek, Francisco Guzmán, Edouard Grave, Myle Ott, Luke Zettlemoyer, and Veselin Stoyanov. 2020. Unsupervised Cross-lingual Representation Learning at Scale. In *Proceedings of the 58th Annual Meeting of the Association for Computational Linguistics*, pages 8440–8451.

DeepSeek-AI. 2026. DeepSeek-V4: Towards Highly Efficient Million-Token Context Intelligence.

Javid Ebrahimi, Anyi Rao, Daniel Lowd, and Dejing Dou. 2018. HotFlip: White-Box Adversarial Examples for Text Classification. In *Proceedings of the 56th Annual Meeting of the Association for Computational Linguistics (Volume 2: Short Papers)*, pages 31–36, Melbourne, Australia.

Mara Finkelstein, Isaac Caswell, Tobias Domhan, Jan-Thorsten Peter, Juraj Juraska, Parker Riley, Daniel Deutsch, Geza Kovacs, Cole Dilanni, Colin Cherry, Eleftheria Briakou, Elizabeth Nielsen, Jiaming Luo, Kat Black, Ryan Mullins, Sweta Agrawal, Wenda Xu, Erin Kats, Stephane Jaskiewicz, Markus Freitag, and David Vilar. 2026. TranslateGemma Technical Report. arXiv: 2601.09012 [cs.CL].

Google DeepMind. 2025. Gemini 3 Flash Model Card. Accessed: 2026-03-15.

Nuno M. Guerreiro, Ricardo Rei, Daan Stigt, Luisa Coheur, Pierre Colombo, and André F. T. Martins. 2024. xCOMET: Transparent Machine Translation Evaluation through Fine-grained Error Detection. *Transactions of the Association for Computational Linguistics* 12:979–995.

Ari Holtzman, Jan Buys, Li Du, Maxwell Forbes, and Yejin Choi. 2020. The Curious Case of Neural Text Degeneration. In *International Conference on Learning Representations (ICLR)*.

Pierre Isabelle, Colin Cherry, and George Foster. 2017. A Challenge Set Approach to Evaluating Machine Translation. In *Proceedings of the 2017 Conference on Empirical Methods in Natural Language Processing*, pages 2486–2496.

Juraj Juraska, Daniel Deutsch, Mara Finkelstein, and Markus Freitag. 2024. MetricX-24: The Google Submission to the WMT 2024 Metrics Shared Task. In *Proceedings of the Ninth Conference on Machine Translation*, pages 492–504, Miami, Florida, USA.

Katharina Kann, Sascha Rothe, and Katja Filippova. 2018. Sentence-Level Fluency Evaluation: References Help, But Can Be Spared! In *Proceedings* *of the 22nd Conference on Computational Natural Language Learning (CoNLL)*, pages 313–323.

Tom Kocmi, Ekaterina Artemova, Eleftherios Avramidis, Rachel Bawden, Ondřej Bojar, Konstantin Dranch, Anton Dvorkovich, Sergey Dukanov, Mark Fishel, Markus Freitag, Thamme Gowda, Roman Grundkiewicz, Barry Haddow, Marzena Karpinska, Philipp Koehn, Howard Lakougna, Jessica Lundin, Christof Monz, Kenton Murray, Masaaki Nagata, Stefano Perrella, Lorenzo Proietti, Martin Popel, Maja Popović, Parker Riley, Mariya Shmatova, Steinthór Steingrímsson, Lisa Yankovskaya, and Vilém Zouhar. 2025. Findings of the WMT25 General Machine Translation Shared Task: Time to Stop Evaluating on Easy Test Sets. In *Proceedings of the Tenth Conference on Machine Translation*, pages 355–413, Suzhou, China.

Tom Kocmi, Eleftherios Avramidis, Rachel Bawden, Ondřej Bojar, Anton Dvorkovich, Christian Federmann, Mark Fishel, Markus Freitag, Thamme Gowda, Roman Grundkiewicz, Barry Haddow, Marzena Karpinska, Philipp Koehn, Benjamin Marie, Christof Monz, Kenton Murray, Masaaki Nagata, Martin Popel, Maja Popović, Mariya Shmatova, Steinthór Steingrímsson, and Vilém Zouhar. 2024. Findings of the WMT24 General Machine Translation Shared Task: The LLM Era Is Here but MT Is Not Solved Yet. In *Proceedings of the Ninth Conference on Machine Translation*, pages 1–46, Miami, Florida, USA.

Tom Kocmi, Eleftherios Avramidis, Rachel Bawden, Ondřej Bojar, Anton Dvorkovich, Christian Federmann, Mark Fishel, Markus Freitag, Thamme Gowda, Roman Grundkiewicz, Barry Haddow, Philipp Koehn, Benjamin Marie, Christof Monz, Makoto Morishita, Kenton Murray, Masaaki Nagata, Toshiaki Nakazawa, Martin Popel, Maja Popović, Mariya Shmatova, and Jun Suzuki. 2023. Findings of the 2023 Conference on Machine Translation (WMT23): LLMs Are Here but Not Quite There Yet. In *Proceedings of the Eighth Conference on Machine Translation*, pages 1–42, Singapore.

Tom Kocmi, Rachel Bawden, Ondřej Bojar, Anton Dvorkovich, Christian Federmann, Mark Fishel, Thamme Gowda, Yvette Graham, Roman Grundkiewicz, Barry Haddow, Rebecca Knowles, Philipp Koehn, Christof Monz, Makoto Morishita, Masaaki Nagata, Toshiaki Nakazawa, Michal Novák, Martin Popel, and Maja Popović. 2022. Findings of the 2022 Conference on Machine Translation (WMT22). In *Proceedings of the Seventh Conference on Machine Translation (WMT)*.

Kalpesh Krishna, Yixiao Song, Marzena Karpinska, John Wieting, and Mohit Iyyer. 2023. Paraphrasing evades detectors of AI-generated text, but retrieval is an effective defense. In *Advances in Neural Information Processing Systems (NeurIPS)*.

Xiaogeng Liu, Nan Xu, Muhao Chen, and Chaowei Xiao. 2024. AutoDAN: Generating Stealthy Jailbreak Prompts on Aligned Large Language Models. In *The Twelfth International Conference on Learning Representations*.

Vivien Macketanz, Eleftherios Avramidis, Aljoscha Burchardt, and Hans Uszkoreit. 2018. Fine-grained evaluation of German-English Machine Translation based on a Test Suite. In *Proceedings of the Third Conference on Machine Translation: Shared Task Papers*, pages 578–587.

NLLB Team. 2024. Scaling neural machine translation to 200 languages. *Nature* 630:841–846.

Kishore Papineni, Salim Roukos, Todd Ward, and Wei-Jing Zhu. 2002. BLEU: A Method for Automatic Evaluation of Machine Translation. In *Proceedings of the 40th Annual Meeting of the Association for Computational Linguistics*, pages 311–318.

Stefano Perrella, Lorenzo Proietti, Alessandro Scirè, Edoardo Barba, and Roberto Navigli. 2024. Guardians of the Machine Translation Meta-Evaluation: Sentinel Metrics Fall In! In *Proceedings of the 62nd Annual Meeting of the Association for Computational Linguistics (Volume 1: Long Papers)*, pages 16216–16244, Bangkok, Thailand.

José Pombal, Nuno M. Guerreiro, Ricardo Rei, and André F. T. Martins. 2025. Zero-shot Benchmarking: A Framework for Flexible and Scalable Automatic Evaluation of Language Models. In *Conference on Language Modeling (COLM)*.

Matt Post. 2018. A Call for Clarity in Reporting BLEU Scores. In *Proceedings of the Third Conference on Machine Translation: Research Papers*, pages 186–191, Brussels, Belgium.

Lorenzo Proietti, Stefano Perrella, Vilém Zouhar, Roberto Navigli, and Tom Kocmi. 2025. Estimating Machine Translation Difficulty. In *Findings of the Association for Computational Linguistics: EMNLP 2025*, pages 24261–24285, Suzhou, China.

Ricardo Rei, Nuno M Guerreiro, José Pombal, João Alves, Pedro Teixeirinha, Amin Farajian, and André FT Martins. 2025. Tower+: Bridging generality and translation specialization in multilingual llms. *arXiv preprint arXiv:2506.17080*.

Yankun Ren, Jianbin Lin, Siliang Tang, Jun Zhou, Shuang Yang, Yuan Qi, and Xiang Ren. 2020. Generating Natural Language Adversarial Examples on a Large Scale with Generative Models. In *ECAI 2020: 24th European Conference on Artificial Intelligence*, pages 2156–2163.

Sahar Sadrizadeh, Clément Barbier, Ljiljana Dolamic, and Pascal Frossard. 2023. A Relaxed Optimization Approach for Adversarial Attacks against Neural Machine Translation Models. In *Proceedings of the*

*31st European Signal Processing Conference (EUSIPCO 2023)*, pages 436–440, EURASIP.

Robyn Speer. 2022. *rspeer/wordfreq: v3.0*. version V3.0.2.

Prashanth Vijayaraghavan and Deb Roy. 2020. Generating Black-Box Adversarial Examples for Text Classifiers Using a Deep Reinforced Model. In *Machine Learning and Knowledge Discovery in Databases*, pages 711–726.

Alex Wang, Amanpreet Singh, Julian Michael, Felix Hill, Omer Levy, and Samuel R. Bowman. 2019. GLUE: A Multi-Task Benchmark and Analysis Platform for Natural Language Understanding. In *International Conference on Learning Representations*.

Alex Warstadt, Amanpreet Singh, and Samuel R. Bowman. 2019. Neural Network Acceptability Judgments. *Transactions of the Association for Computational Linguistics* 7:625–641.

Wenda Xu, Vilém Zouhar, Parker Riley, Mara Finkelstein, Markus Freitag, and Daniel Deutsch. 2025. Searching for Difficult-to-Translate Test Examples at Scale. *arXiv preprint arXiv:2509.26619*.

Xinghao Yang, Weifeng Liu, Dacheng Tao, and Wei Liu. 2021. BESA: BERT-based Simulated Annealing for Adversarial Text Attacks. In *Proceedings of the Thirtieth International Joint Conference on Artificial Intelligence*, pages 3293–3299.

Hai Zhu, Qinyang Zhao, and Yuren Wu. 2023. BeamAttack: Generating High-quality Textual Adversarial Examples Through Beam Search and Mixed Semantic Spaces. In *Advances in Knowledge Discovery and Data Mining*, pages 454–465, Cham.

Andy Zou, Zifan Wang, Nicholas Carlini, Milad Nasr, J Zico Kolter, and Matt Fredrikson. 2023. Universal and transferable adversarial attacks on aligned language models. *arXiv preprint arXiv:2307.15043*.

Vilém Zouhar and Tom Kocmi. 2026. Pearmut: Human Evaluation of Translation Made Trivial. arXiv: 2601.02933 [cs.CL].

Vilém Zouhar, Wenda Xu, Parker Riley, Juraj Juraska, Mara Finkelstein, Markus Freitag, and Daniel Deutsch. 2025. Generating Difficult-to-Translate Texts. *arXiv preprint arXiv:2509.26592*.

## A Additional Results

We present additional results from the automatic evaluation.

### A.1 xCOMET

| | NLLB-200-3.3B | Tower-Plus-72B | TranslateGemma-27b | Gemini-3-Flash | DeepSeek-V4-Flash |
|---|---|---|---|---|---|
| **EN-CS** | | | | | |
| Base text | 0.89 ±0.02 | 0.94 ±0.01 | 0.95 ±0.01 | 0.95 ±0.01 | 0.94 ±0.01 |
| Random Subword | 0.66 ±0.04 | 0.81 ±0.03 | 0.84 ±0.02 | 0.84 ±0.02 | 0.81 ±0.02 |
| Random Word (Common) | 0.60 ±0.04 | 0.75 ±0.03 | 0.80 ±0.02 | 0.78 ±0.02 | 0.77 ±0.02 |
| Random Word (Rare) | 0.64 ±0.04 | 0.81 ±0.02 | 0.83 ±0.02 | 0.83 ±0.02 | 0.80 ±0.02 |
| Paraphrasing | 0.78 ±0.04 | 0.90 ±0.02 | 0.92 ±0.01 | 0.91 ±0.02 | 0.90 ±0.02 |
| Zeroshot (Qwen 2.5 72B) | 0.74 ±0.04 | 0.88 ±0.02 | 0.91 ±0.02 | 0.89 ±0.02 | 0.89 ±0.02 |
| Zeroshot (DeepSeek-V4-Flash) | 0.72 ±0.04 | 0.87 ±0.02 | 0.90 ±0.02 | 0.88 ±0.02 | 0.87 ±0.02 |
| ATO-Direct | 0.78 ±0.03 | 0.86 ±0.02 | 0.90 ±0.02 | 0.89 ±0.02 | 0.87 ±0.02 |
| ATO-TwoPhase | 0.73 ±0.03 | 0.82 ±0.02 | 0.85 ±0.02 | 0.81 ±0.03 | 0.80 ±0.03 |
| **EN-DE** | | | | | |
| Base text | 0.94 ±0.02 | 0.97 ±0.01 | 0.97 ±0.01 | 0.97 ±0.01 | 0.97 ±0.01 |
| Random Subword | 0.77 ±0.04 | 0.91 ±0.01 | 0.91 ±0.01 | 0.91 ±0.01 | 0.91 ±0.01 |
| Random Word (Common) | 0.73 ±0.04 | 0.89 ±0.01 | 0.89 ±0.01 | 0.89 ±0.01 | 0.88 ±0.01 |
| Random Word (Rare) | 0.76 ±0.04 | 0.90 ±0.01 | 0.91 ±0.01 | 0.91 ±0.01 | 0.90 ±0.01 |
| Paraphrasing | 0.86 ±0.03 | 0.95 ±0.01 | 0.95 ±0.01 | 0.95 ±0.01 | 0.95 ±0.01 |
| Zeroshot (Qwen 2.5 72B) | 0.86 ±0.03 | 0.93 ±0.01 | 0.94 ±0.01 | 0.94 ±0.01 | 0.93 ±0.01 |
| Zeroshot (DeepSeek-V4-Flash) | 0.85 ±0.03 | 0.93 ±0.01 | 0.94 ±0.01 | 0.93 ±0.01 | 0.92 ±0.01 |
| ATO-Direct | 0.87 ±0.02 | 0.95 ±0.01 | 0.95 ±0.01 | 0.94 ±0.01 | 0.94 ±0.01 |
| ATO-TwoPhase | 0.85 ±0.03 | 0.92 ±0.01 | 0.92 ±0.01 | 0.91 ±0.01 | 0.92 ±0.01 |
| **EN-IS** | | | | | |
| Base text | 0.83 ±0.03 | 0.90 ±0.02 | 0.89 ±0.02 | 0.92 ±0.01 | 0.91 ±0.01 |
| Random Subword | 0.58 ±0.04 | 0.77 ±0.03 | 0.78 ±0.02 | 0.44 ±0.04 | 0.79 ±0.02 |
| Random Word (Common) | 0.51 ±0.03 | 0.75 ±0.02 | 0.72 ±0.03 | 0.79 ±0.02 | 0.76 ±0.02 |
| Random Word (Rare) | 0.56 ±0.04 | 0.78 ±0.02 | 0.76 ±0.03 | 0.81 ±0.02 | 0.76 ±0.03 |
| Paraphrasing | 0.74 ±0.04 | 0.86 ±0.02 | 0.85 ±0.02 | 0.89 ±0.02 | 0.88 ±0.02 |
| Zeroshot (Qwen 2.5 72B) | 0.68 ±0.04 | 0.83 ±0.02 | 0.83 ±0.02 | 0.87 ±0.02 | 0.85 ±0.02 |
| Zeroshot (DeepSeek-V4-Flash) | 0.67 ±0.04 | 0.82 ±0.02 | 0.83 ±0.02 | 0.85 ±0.02 | 0.83 ±0.02 |
| ATO-Direct | 0.70 ±0.03 | 0.84 ±0.02 | 0.82 ±0.02 | 0.86 ±0.02 | 0.83 ±0.02 |
| ATO-TwoPhase | 0.68 ±0.03 | 0.81 ±0.02 | 0.80 ±0.02 | 0.81 ±0.02 | 0.80 ±0.02 |
| **EN-RU** | | | | | |
| Base text | 0.91 ±0.02 | 0.95 ±0.01 | 0.96 ±0.01 | 0.95 ±0.01 | 0.95 ±0.01 |
| Random Subword | 0.70 ±0.03 | 0.84 ±0.02 | 0.85 ±0.02 | 0.86 ±0.02 | 0.85 ±0.02 |
| Random Word (Common) | 0.67 ±0.03 | 0.82 ±0.02 | 0.83 ±0.02 | 0.82 ±0.02 | 0.81 ±0.02 |
| Random Word (Rare) | 0.67 ±0.03 | 0.85 ±0.02 | 0.84 ±0.02 | 0.82 ±0.02 | 0.83 ±0.02 |
| Paraphrasing | 0.80 ±0.04 | 0.92 ±0.02 | 0.93 ±0.01 | 0.92 ±0.01 | 0.91 ±0.02 |
| Zeroshot (Qwen 2.5 72B) | 0.77 ±0.04 | 0.89 ±0.02 | 0.91 ±0.02 | 0.90 ±0.02 | 0.89 ±0.02 |
| Zeroshot (DeepSeek-V4-Flash) | 0.76 ±0.03 | 0.89 ±0.02 | 0.90 ±0.02 | 0.89 ±0.02 | 0.88 ±0.02 |
| ATO-Direct | 0.81 ±0.02 | 0.90 ±0.01 | 0.90 ±0.02 | 0.88 ±0.02 | 0.88 ±0.02 |
| ATO-TwoPhase | 0.77 ±0.03 | 0.85 ±0.02 | 0.85 ±0.02 | 0.83 ±0.02 | 0.83 ±0.02 |
| **EN-ES** | | | | | |
| Base text | 0.92 ±0.02 | 0.95 ±0.01 | 0.95 ±0.01 | 0.94 ±0.01 | 0.94 ±0.01 |
| Random Subword | 0.71 ±0.03 | 0.86 ±0.02 | 0.86 ±0.02 | 0.84 ±0.02 | 0.85 ±0.02 |
| Random Word (Common) | 0.67 ±0.03 | 0.82 ±0.02 | 0.83 ±0.02 | 0.79 ±0.02 | 0.82 ±0.02 |
| Random Word (Rare) | 0.70 ±0.03 | 0.84 ±0.02 | 0.85 ±0.02 | 0.78 ±0.02 | 0.84 ±0.02 |
| Paraphrasing | 0.86 ±0.03 | 0.92 ±0.01 | 0.93 ±0.01 | 0.92 ±0.01 | 0.91 ±0.01 |
| Zeroshot (Qwen 2.5 72B) | 0.82 ±0.03 | 0.90 ±0.01 | 0.92 ±0.01 | 0.89 ±0.02 | 0.89 ±0.02 |
| Zeroshot (DeepSeek-V4-Flash) | 0.81 ±0.03 | 0.89 ±0.02 | 0.90 ±0.01 | 0.89 ±0.02 | 0.88 ±0.02 |
| ATO-Direct | 0.82 ±0.02 | 0.90 ±0.01 | 0.90 ±0.02 | 0.86 ±0.02 | 0.88 ±0.01 |
| ATO-TwoPhase | 0.80 ±0.02 | 0.86 ±0.02 | 0.87 ±0.02 | 0.85 ±0.02 | 0.84 ±0.02 |

Table 7: Per-language, per-model xCOMET scores with 95% confidence intervals across all augmentation methods.

## A.2 MetricX

| | NLLB-200-3.3B | Tower-Plus-72B | TranslateGemma-27b | Gemini-3-Flash | DeepSeek-V4-Flash |
|---|---|---|---|---|---|
| **EN-CS** | | | | | |
| Base text | 4.43 ±0.52 | 2.91 ±0.26 | 2.43 ±0.20 | 2.72 ±0.24 | 2.95 ±0.26 |
| Random Subword | 8.97 ±0.72 | 5.85 ±0.43 | 4.55 ±0.30 | 5.98 ±0.44 | 5.97 ±0.42 |
| Random Word (Common) | 9.32 ±0.71 | 5.64 ±0.45 | 4.31 ±0.32 | 5.73 ±0.46 | 5.81 ±0.45 |
| Random Word (Rare) | 9.11 ±0.74 | 5.95 ±0.43 | 4.72 ±0.34 | 5.94 ±0.45 | 6.21 ±0.44 |
| Paraphrasing | 6.20 ±0.75 | 3.28 ±0.26 | 2.48 ±0.18 | 3.00 ±0.24 | 3.14 ±0.24 |
| Zeroshot (Qwen 2.5 72B) | 7.05 ±0.80 | 3.59 ±0.30 | 2.77 ±0.21 | 3.16 ±0.24 | 3.61 ±0.31 |
| Zeroshot (DeepSeek-V4-Flash) | 7.28 ±0.77 | 3.78 ±0.34 | 2.89 ±0.24 | 3.32 ±0.28 | 3.84 ±0.35 |
| ATO-Direct | 6.46 ±0.59 | 4.15 ±0.31 | 3.21 ±0.24 | 4.30 ±0.35 | 4.26 ±0.35 |
| ATO-TwoPhase | 7.53 ±0.66 | 5.21 ±0.42 | 3.95 ±0.35 | 5.78 ±0.50 | 5.68 ±0.47 |
| **EN-DE** | | | | | |
| Base text | 1.65 ±0.43 | 0.67 ±0.10 | 0.55 ±0.09 | 0.75 ±0.15 | 0.84 ±0.16 |
| Random Subword | 6.00 ±0.79 | 2.62 ±0.24 | 2.30 ±0.22 | 3.21 ±0.32 | 2.77 ±0.26 |
| Random Word (Common) | 6.18 ±0.82 | 2.50 ±0.27 | 1.87 ±0.18 | 2.66 ±0.28 | 2.70 ±0.25 |
| Random Word (Rare) | 6.02 ±0.82 | 2.84 ±0.26 | 2.32 ±0.21 | 2.86 ±0.25 | 2.86 ±0.24 |
| Paraphrasing | 3.17 ±0.67 | 0.91 ±0.13 | 0.78 ±0.11 | 0.90 ±0.13 | 0.97 ±0.14 |
| Zeroshot (Qwen 2.5 72B) | 3.10 ±0.62 | 1.08 ±0.13 | 0.87 ±0.12 | 1.01 ±0.12 | 1.13 ±0.14 |
| Zeroshot (DeepSeek-V4-Flash) | 3.21 ±0.58 | 1.15 ±0.15 | 0.88 ±0.12 | 1.11 ±0.15 | 1.27 ±0.16 |
| ATO-Direct | 3.32 ±0.60 | 1.32 ±0.15 | 1.10 ±0.14 | 1.61 ±0.19 | 1.64 ±0.20 |
| ATO-TwoPhase | 3.75 ±0.60 | 2.08 ±0.27 | 1.42 ±0.19 | 2.40 ±0.33 | 2.29 ±0.30 |
| **EN-IS** | | | | | |
| Base text | 5.20 ±0.55 | 3.49 ±0.37 | 3.16 ±0.32 | 2.75 ±0.28 | 3.44 ±0.35 |
| Random Subword | 10.19 ±0.72 | 5.94 ±0.46 | 5.26 ±0.38 | 5.52 ±0.43 | 6.13 ±0.45 |
| Random Word (Common) | 10.67 ±0.74 | 5.38 ±0.38 | 4.94 ±0.38 | 5.22 ±0.40 | 6.06 ±0.46 |
| Random Word (Rare) | 10.52 ±0.74 | 5.87 ±0.43 | 5.15 ±0.38 | 5.53 ±0.42 | 6.12 ±0.44 |
| Paraphrasing | 6.79 ±0.76 | 4.00 ±0.37 | 3.67 ±0.33 | 3.04 ±0.27 | 3.51 ±0.31 |
| Zeroshot (Qwen 2.5 72B) | 7.82 ±0.77 | 3.88 ±0.33 | 3.75 ±0.33 | 3.16 ±0.28 | 4.13 ±0.39 |
| Zeroshot (DeepSeek-V4-Flash) | 8.02 ±0.77 | 4.12 ±0.38 | 3.82 ±0.32 | 3.35 ±0.29 | 4.00 ±0.35 |
| ATO-Direct | 7.29 ±0.58 | 4.74 ±0.40 | 4.27 ±0.38 | 4.55 ±0.40 | 5.06 ±0.42 |
| ATO-TwoPhase | 8.50 ±0.67 | 5.55 ±0.48 | 4.61 ±0.40 | 5.75 ±0.52 | 6.51 ±0.56 |
| **EN-RU** | | | | | |
| Base text | 2.66 ±0.47 | 1.33 ±0.20 | 0.98 ±0.16 | 1.29 ±0.19 | 1.44 ±0.23 |
| Random Subword | 7.27 ±0.72 | 3.68 ±0.33 | 2.87 ±0.27 | 3.88 ±0.34 | 4.03 ±0.35 |
| Random Word (Common) | 7.51 ±0.71 | 3.44 ±0.34 | 2.63 ±0.27 | 3.93 ±0.42 | 4.09 ±0.41 |
| Random Word (Rare) | 7.28 ±0.69 | 3.72 ±0.31 | 2.92 ±0.25 | 4.95 ±0.46 | 4.21 ±0.38 |
| Paraphrasing | 4.82 ±0.79 | 1.74 ±0.23 | 1.32 ±0.18 | 1.63 ±0.20 | 1.75 ±0.21 |
| Zeroshot (Qwen 2.5 72B) | 5.30 ±0.78 | 1.92 ±0.23 | 1.49 ±0.19 | 1.77 ±0.21 | 2.12 ±0.24 |
| Zeroshot (DeepSeek-V4-Flash) | 5.45 ±0.77 | 2.01 ±0.28 | 1.53 ±0.20 | 1.95 ±0.25 | 2.11 ±0.26 |
| ATO-Direct | 4.13 ±0.48 | 2.22 ±0.28 | 1.77 ±0.23 | 2.57 ±0.31 | 2.54 ±0.29 |
| ATO-TwoPhase | 5.09 ±0.54 | 3.25 ±0.37 | 2.40 ±0.29 | 3.78 ±0.39 | 3.83 ±0.39 |
| **EN-ES** | | | | | |
| Base text | 2.76 ±0.44 | 1.78 ±0.17 | 1.48 ±0.15 | 1.77 ±0.16 | 1.90 ±0.18 |
| Random Subword | 7.49 ±0.70 | 4.18 ±0.32 | 3.59 ±0.27 | 4.57 ±0.36 | 4.35 ±0.32 |
| Random Word (Common) | 7.75 ±0.70 | 4.01 ±0.32 | 3.32 ±0.28 | 4.71 ±0.41 | 4.31 ±0.35 |
| Random Word (Rare) | 7.62 ±0.72 | 4.31 ±0.33 | 3.68 ±0.27 | 6.71 ±0.48 | 4.74 ±0.36 |
| Paraphrasing | 3.71 ±0.58 | 1.99 ±0.18 | 1.72 ±0.14 | 2.00 ±0.17 | 2.10 ±0.20 |
| Zeroshot (Qwen 2.5 72B) | 4.13 ±0.60 | 2.24 ±0.18 | 1.85 ±0.16 | 2.26 ±0.20 | 2.38 ±0.20 |
| Zeroshot (DeepSeek-V4-Flash) | 4.49 ±0.60 | 2.39 ±0.24 | 1.97 ±0.18 | 2.33 ±0.22 | 2.53 ±0.27 |
| ATO-Direct | 4.59 ±0.49 | 2.84 ±0.25 | 2.19 ±0.19 | 3.77 ±0.31 | 3.35 ±0.30 |
| ATO-TwoPhase | 5.16 ±0.52 | 3.91 ±0.38 | 3.02 ±0.35 | 4.46 ±0.44 | 4.26 ±0.40 |

Table 8: Per-language, per-model MetricX scores with 95% confidence intervals across all augmentation methods.

## A.3 Data Metrics

A potential concern is vocabulary collapse: the optimizer might converge on a small set of tokens that Sentinel considers difficult and insert them into every sequence. To test for this, we measure word diversity: the number of unique words across all sequences, normalized by total word count. We also report average word length (in characters) and average word count (by whitespace). Results are shown in Table 9. Word count increases steadily across methods, consistent with the intuition that longer texts are harder to translate. Word diversity remains stable, indicating that the optimizer does not degenerate into repeated insertion of a fixed token set. Word length shows no clear trend.

| Method | Word Count | Avg Word Length | Word Diversity |
|---|---|---|---|
| Base text | 17.25 ±1.59 | 5.17 ±0.14 | 0.588 |
| Random Subword | 17.00 ±1.57 | 5.43 ±0.16 | 0.649 |
| Random Word (Common) | 17.25 ±1.59 | 5.46 ±0.15 | 0.647 |
| Random Word (Rare) | 17.25 ±1.59 | 5.24 ±0.13 | 0.658 |
| Paraphrasing | 17.95 ±1.68 | 5.01 ±0.14 | 0.556 |
| Zeroshot (Qwen 2.5 72B) | 17.25 ±1.59 | 5.23 ±0.14 | 0.605 |
| Zeroshot (DeepSeek-V4-Flash) | 17.45 ±1.61 | 5.13 ±0.13 | 0.598 |
| ATO-Direct | 17.26 ±1.59 | 5.29 ±0.15 | 0.612 |
| ATO-TwoPhase | 17.15 ±1.58 | 5.26 ±0.17 | 0.604 |

Table 9: Analysis of word quality metrics including length and diversity across the different generation methods.

# B Model Prompts & Details

## B.1 Translation Prompt

To use Gemini-3-Flash and DeepSeek-V4-Flash for translation, we used the prompt stated in Figure 3. The strings for translation have been batched by 10 to increase throughput. The model was accessed through the official Google API under the name `models/gemini-3-flash-preview` on 19-03-2026. DeepSeek-V4-Flash has been used through OpenRouter API under the name `deepseek/deepseek-v4-flash` on 06-05-2026.

```
You are a professional translator. Translate the following list of strings into the target language.
Maintain the exact order of the list.
Return ONLY a valid Python-style list of strings with the translations.
Do not include explanations, code blocks, or markdown. CRITICAL: Do not skip any sequences even if they are difficult to translate!
Source language: SRCLANGUAGE
Target language: TGTLANGUAGE
```

Figure 3: Translation Prompt

## B.2 Zeroshot Baseline Prompt

To use Qwen 2.5 72B and DeepSeek-V4-Flash as zero-shot replacement baselines, we used the prompt in Figure 4. DeepSeek-V4-Flash was accessed through the OpenRouter API as `deepseek/deepseek-v4-flash` on 06-05-2026.

```
You are a linguistics expert specializing in translation difficulty.
I will give you an English text. Pick EXACTLY {num_replacements} word(s) to replace with alternatives that make the text harder to translate into German. Return your answer as a JSON object with exactly {num_replacements} entries:
{"1": {"original": "word1", "replacement": "new_word1"}, "2": {"original": "word2", "replacement": "new_word2"}}
Rules:
• Pick exactly {num_replacements} word(s) to replace – no more, no less.
• Each replacement must be a single word.
• Choose words that are ambiguous, idiomatic, or culturally specific.
• Do NOT use German words.
• Return ONLY the JSON object, nothing else.
Text: {text}
```

Figure 4: Zeroshot Prompt

## B.3 DIPPER Paraphrasing Baseline Details

DIPPER takes a text as input and outputs a rephrased version of that text. It also accepts a lexical diversity parameter controlling how aggressively words are changed and an order diversity parameter controlling word reordering. We fixed order diversity to zero, since our optimization does not rearrange words, and selected a lexical diversity of 20 by matching the edit distance distribution of DIPPER's output to that of our two-phase method across the full range (0–100).

### B.4 Random Replacement Baseline Details

We replace either whole words from the vocabulary in Section 3.3 or subwords from Section 3.4, replacing exactly 2 words or subwords per seed sentence. For whole-word tokens we distinguish between common and rare using Zipf word frequency from the `wordfreq` library (Speer, 2022). We require Zipf frequency greater than 2.5 to filter out extremely rare words, then split the remainder at the median into common (Zipf $\geq 4.09$) and rare ($2.5 <$ Zipf $< 4.09$) buckets.

## C Human Evaluation

### C.1 Grammaticality and Plausibility

Rate the text below on two dimensions:
Grammaticality (1–5):
How grammatically well-formed is the text?
1 = Completely ungrammatical — severe errors that make the text very hard/impossible to understand.
2 = Major grammatical problems — understandable, but errors clearly disrupt structure or fluency.
3 = Noticeable issues — errors present but do not seriously impact comprehension.
4 = Minor imperfections — almost fully grammatical, only very small issues (e.g. a typo).
5 = Fully grammatical — no noticeable errors; the text is well-formed.
Plausibility (1–5):
How likely is it that this text would appear in real online content?
1 = Implausible — this text would not appear in any realistic context.
2 = Unlikely — would rarely occur, even though it is not impossible.
3 = Moderately plausible — could appear, but would be somewhat unusual.
4 = Quite plausible — would commonly make sense in online content.
5 = Highly plausible — would very naturally appear as part of real online content.“

Figure 5: Instructions for Rating Grammaticality and Plausibility

### C.2 Translation Quality

Read the original English text and its machine translation carefully.
Translation Quality (1–5):
How well does the translation convey the original meaning?
1 = Poor — the translation is inaccurate, unintelligible, or fails to convey the original meaning.
2 = Fair — significant errors or awkward phrasing that affect clarity.
3 = Good — mostly accurate and understandable, with minor issues.
4 = Very Good — accurate and fluent, with only negligible errors.
5 = Excellent — accurate, fluent, and natural-sounding.

Figure 6: Instructions for Rating Translation Quality

### C.3 Inter-Annotator Agreement

| Metric | $\alpha$ absolute | $\alpha$ relative |
|---|---|---|
| Grammaticality | 0.18 | 0.41 |
| Plausibility | 0.28 | 0.41 |
| Translation quality | 0.79 | 0.51 |

Table 10: Krippendorff's $\alpha$ on human evaluation metrics. For relative $\alpha$, each rater's ratings are first $z$-normalised within rater (centered on that rater's mean, scaled by their standard deviation), and Krippendorff's $\alpha$ is then computed at the interval level on the resulting (rater $\times$ item) matrix.

Although the absolute $\alpha$ varies across metrics (0.18–0.79), the relative $\alpha$ range of 0.41–0.51 suggests per-rater scale bias causes absolute variation rather than disagreement on the relative ordering of items.

## C.4 Evaluation interface

We use Pearmut (Zouhar and Kocmi, 2026), an open-source translation evaluation tool. The evaluations are thus reproducible given the data to be evaluated.

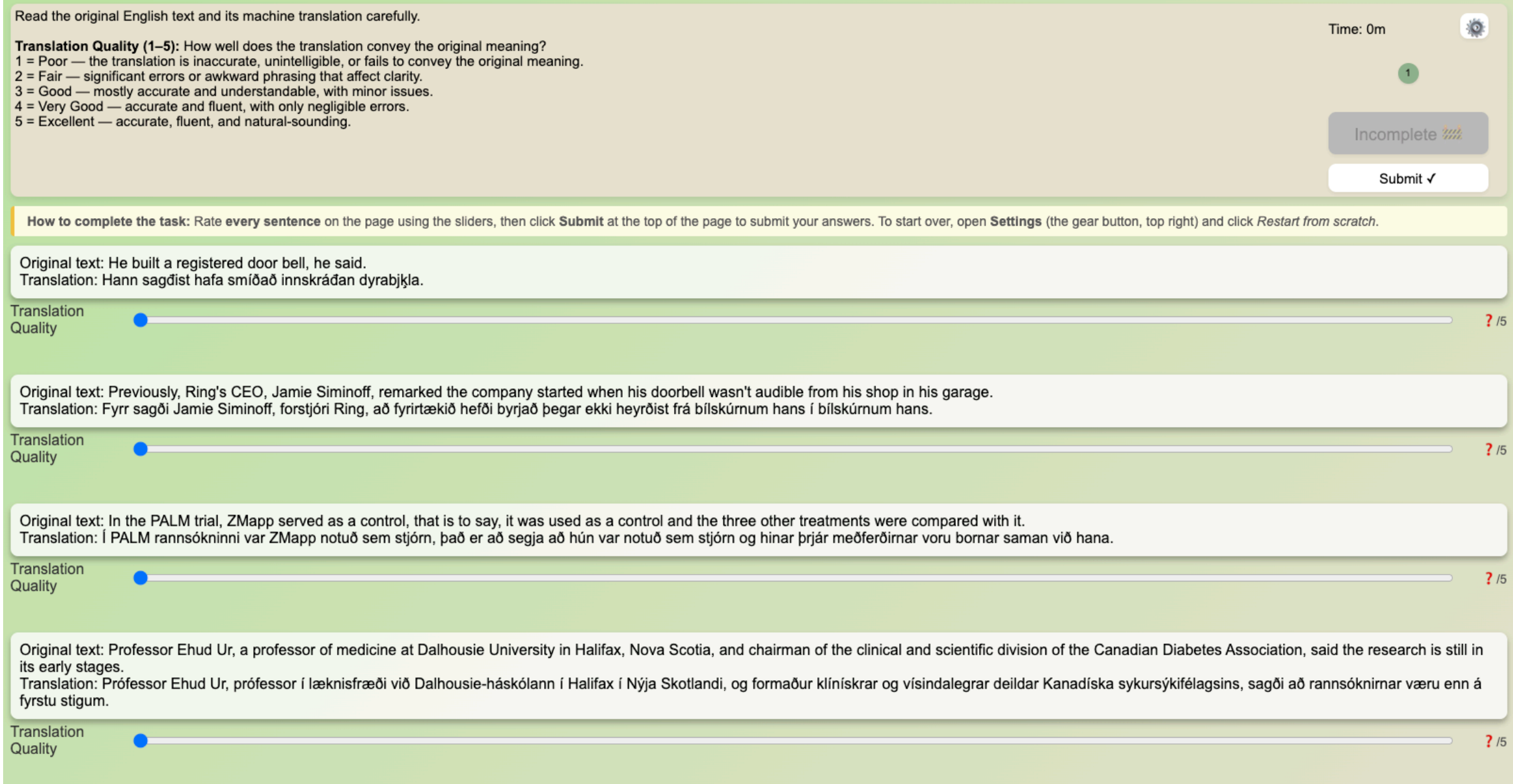

Figure 7: Translation quality evaluation interface. Pictured language is Icelandic, and we had a separate interface for each of the other languages.

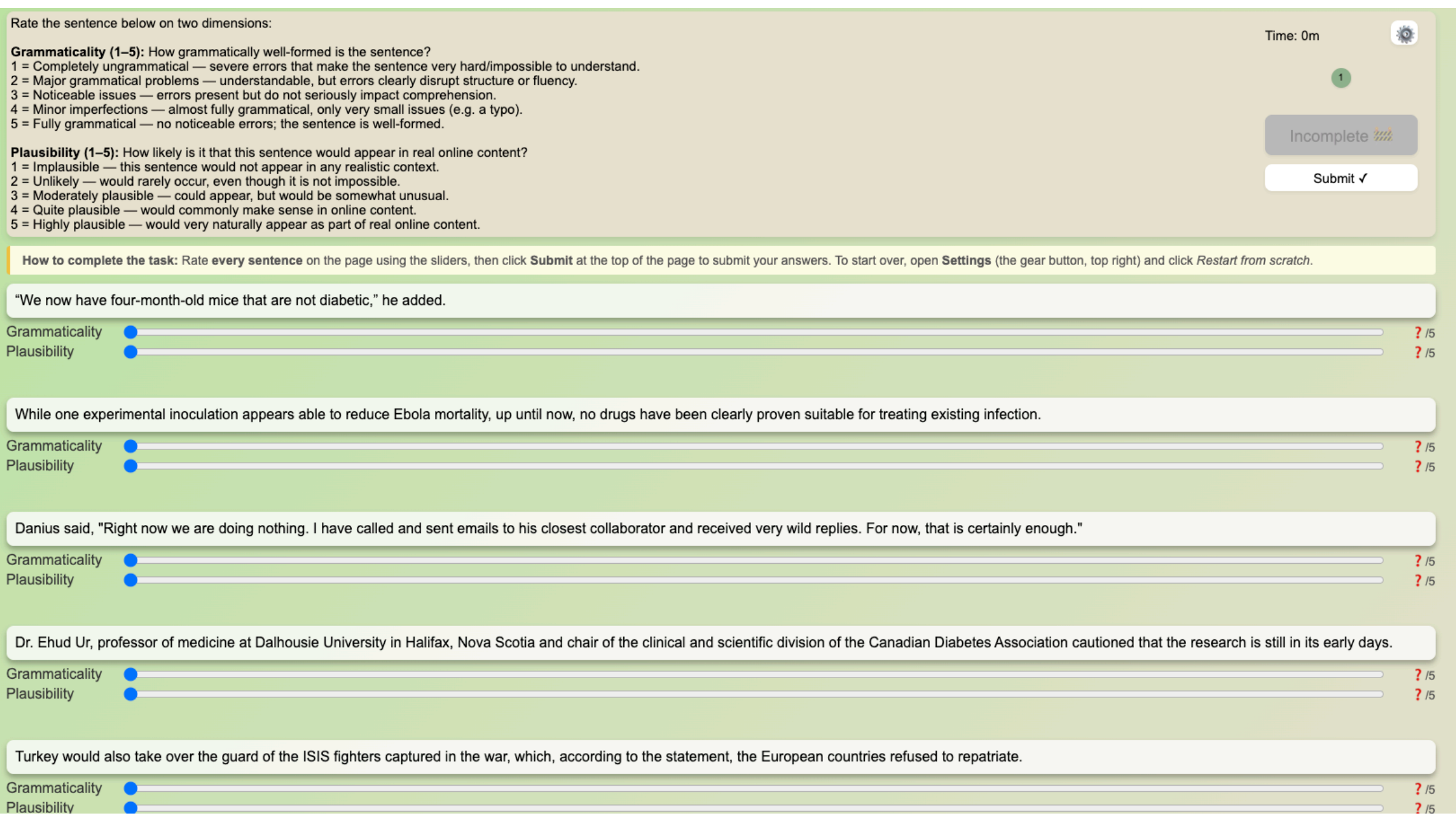

Figure 8: English text quality evaluation interface.